\documentclass[final,5p,times,twocolumn]{modified} 

\usepackage{mathtools,amsmath,amssymb}
\usepackage[hyphens]{url}
\usepackage{stmaryrd}
\usepackage{paralist}
\usepackage{graphicx}
\usepackage{rotating}
\usepackage{subcaption}
\usepackage{boldline}
\usepackage{framed,enumerate}

\usepackage[T1]{fontenc}

\usepackage{multirow}

\usepackage{macros}

\usepackage{parskip,textcomp}

\usepackage{color}

\begin{document}

\begin{frontmatter}

\title{Towards Human-Compatible XAI:\\ Explaining Data Differentials with Concept Induction over Background Knowledge\tnoteref{ack}}
\tnotetext[ack]{This research was, in part, funded by the U.S. Government via contract FA86492099045. The views and conclusions contained in this document are those of the authors and should not be interpreted as representing the official policies, either expressed or implied, of the U.S. Government.}

\author[kairos]{Cara Widmer}
\ead{cara@kairosresearch.com}

\author[hart]{Md Kamruzzaman Sarker}
\ead{sarker@hartford.edu}

\author[kairos]{Srikanth Nadella}
\ead{srikanth@kairosresearch.com}

\author[kairos]{Joshua Fiechter}
\ead{josh@kairosresearch.com}

\author[kairos,wright]{Ion Juvina}
\ead{ion.juvina@wright.edu}

\author[kairos]{Brandon Minnery}
\ead{brad@kairosresearch.com}

\author[ks]{Pascal Hitzler}
\ead{hitzler@ksu.edu}

\author[ks]{Joshua Schwartz}
\ead{jschwartz427@ksu.edu}

\author[kairos,wright]{Michael Raymer}
\ead{michael.raymer@wright.edu}

\address[kairos]{Kairos Research, LLC}
\address[hart]{University of Hartford, USA}
\address[ks]{Kansas State University, USA}
\address[wright]{Wright State University}

\begin{abstract}
Concept induction, which is based on formal logical reasoning over description logics, has been used in ontology engineering in order to create ontology (TBox) axioms from the base data (ABox) graph. In this paper, we show that it can also be used to explain data differentials, for example in the context of Explainable AI (XAI), and we show that it can in fact be done in a way that is meaningful to a human observer. Our approach utilizes a large class hierarchy, curated from the Wikipedia category hierarchy, as background knowledge. 
\end{abstract}


\end{frontmatter}

\section{Introduction}
\label{sec:intro}

Lack of explainability and understandability are pervasive problems in modern artificial intelligence. Many modern machine learning (ML) methods in particular involve little to no human intervention once training data has been provided, and the resulting models are often "black boxes" whose internal representations and decision mechanisms remain opaque to human users \cite{castelvecchi}. In order for human operators to be able to trust an ML algorithm's outputs, make actionable decisions based on those outputs, and detect and correct biases, it is important that the algorithm's reasoning processes be understandable. Thus there has been an increased focus on explainable artificial intelligence (XAI) in recent years \cite{arrieta}.

Two major approaches for improving explainability are implementing transparency in the ML models themselves and inferring explanations via post-hoc techniques \cite{arrieta}. For example, transparency can be directly introduced to ML models by incorporating simulatability (i.e., developing the model in such a way that the model can be simulated in the thoughts of a human analyzing the model output) \cite{tibshirani} or decomposability (i.e., enabling understandability of each part of the model individually) \cite{lou}.

In contrast, post-hoc explainability techniques do not directly alter a model to make it more explainable but rather infer an explanation based on analysis of the model’s behaviors. Examples of post-hoc techniques include generating human-understandable text explanations and visualizations of a model's behavior, as well as analyses that compute quantitative measures of the influence of various input variables on the model's  output. While post-hoc methods explain a model’s decisions indirectly, they avoid the performance trade-offs that often accompany attempts to engineer transparency directly into the model \cite{dosilovic}.

In the current work we implement a different post-hoc explainability strategy by applying concept induction \cite{conceptLearning} to provide human-interpretable explanations of machine learning classifications. We demonstrate that concept induction, together with a suitable class hierarchy as background knowledge, can be used to generate explanations for data differences that are meaningful for a human observer. A class hierarchy consisting of a curated form of the Wikipedia category hierarchy \cite{DBLP:conf/kgswc/SarkerSHZNMJRA20} was used as background ontology for concept induction, and the ECII concept induction system \cite{DBLP:conf/aaai/SarkerH19} was used for explanation generation. We report on experiments that we have conducted with Amazon Mechanical Turk to assess how meaningful the generated explanations are for humans. The first experiment assesses the quality of explanations generated by concept induction in a specific setting related to scene classification, and supports our hypothesis that these explanations are judged by human participants to be more accurate than semi-random explanation, but less accurate than human-authored explanations. The second experiment shows that concept induction can be used to generate explanations for errors made by a ML classifier; the study also shows that these explanations are inferior to human-generated explanations.

Data and other resources for the work reported herein are available from \url{https://osf.io/xjkcw/}.

The paper is structured as follows. In Section \ref{sec:conceptInduction} we review the concept induction technique and discuss the background knowledge and data mapping process underlying our work. In Sections \ref{sec:firstExp} and \ref{sec:secondExp} we detail our two experiments. In Section \ref{sec:related} we present related efforts, and in Section \ref{sec:conc} we discuss future work and conclude. 

\section{Concept Induction and Utilized Background Knowledge}
\label{sec:conceptInduction}

Concept Induction \cite{conceptLearning} in its general form can be defined as follows. As background knowledge we assume that we have a given description logic ontology\footnote{See \cite{FOST} for background on description logics and their relation to ontologies, in particular the Web Ontology Language OWL \cite{owl2-primer}.} $O$ with TBox $T$ and ABox $A$. We further assume that we are given two sets $P$ and $N$ each consisting of individuals from $O$: $P$ is called the set of \emph{positive} examples and $N$ is called the set of \emph{negative} examples. The concept induction task then consists of identifying complex description logic expressions $C$ such that $O\models C(p)$ for all $p\in P$ and $O\not\models C(n)$ for all $n\in N$. If no such $C$ exists, then a concept induction system is expected to return approximate solutions, together with some accuracy value.\footnote{There are different ways to assess accuracy, including precision, recall, and F1 scores.}

Generally speaking, concept induction is computationally very expensive. Known provably correct algorithms, such as the one underlying the well-known DL-Learner system \cite{LehmannH10,BuhmannLW16}, proceed by \emph{concept refinement}, which is an adaptation from inductive logic programming \cite{DBLP:journals/ngc/Muggleton91}: A candidate concept expression $C_0$ is assessed as to whether $O\models C(p)$ for all $p\in P$ and $O\not\models C(n)$ for all $n\in N$ hold, and if not, then several candidate \emph{refinements} of $C_0$, meaning concept expressions that are somewhat more or less general than $C_0$, are assessed, and the best of these becomes the new candidate concept expression $C_1$. Then this process is repeated until either a perfect solution is found or some termination criterion is met. Note that for each candidate concept, several calls to a description logic reasoner have to be made for assessment, and in particular with large sets $P,S$ and a large expressive ontology $O$, each of these reasoner calls can take considerable time.

In this paper, we are thus basing our analysis on a heuristic algorithm and system, called ECII (Efficient Concept Induction from Individuals) \cite{DBLP:conf/aaai/SarkerH19}. The heuristic underlying ECII rests primarily on (a) using only a class hierarchy as ontology, rather than a more complex ontology, (b) a limited search space for candidate concept expressions, and (c) a priori materialization of logical consequences. With ECII, it is possible to perform concept induction over data sets that cannot otherwise be dealt with, and experiments have shown that output accuracy is still rather high \cite{DBLP:conf/aaai/SarkerH19}.

Previous uses of concept induction have been in the context of ontology engineering \cite{DBLP:conf/semweb/LehmannB10,BuhmannLW16,DBLP:journals/semweb/Paulheim17}. We have previously proposed concept induction for explainability, and preliminary studies have been published \cite{SarkerXDRH17,DBLP:conf/kgswc/SarkerSHZNMJRA20}. In particular, \cite{DBLP:conf/aaai/SarkerH19,DBLP:conf/kgswc/SarkerSHZNMJRA20} presented preparatory methods that enabled the work we discuss in this paper: the ECII system \cite{DBLP:conf/aaai/SarkerH19}, and our curated Wikipedia category hierarchy \cite{DBLP:conf/kgswc/SarkerSHZNMJRA20}. 

The latter was produced by first retrieving the complete category hierarchy from Wikipedia. Since it is crowdsourced, it actually contains cycles which, if interpreted as a proper class hierarchy (i.e., in the sense of rdfs:subClassOf), would make parts of it collapse. We thus devised an algorithm for breaking these cycles heuristically. The resulting class hierarchy consists of 1,860,342 concepts. In addition, 6,079,748 individuals -- which correspond to Wikipedia page URLs -- are assigned to the classes according to the categorizations found on Wikipedia. 

Using the Wikipedia category hierarchy also enabled us to establish an easy mapping process of examples into the hierarchy: Text labels (more details about the dataset we used can be found in Sections \ref{sec:firstExp} and \ref{sec:secondExp}) were mapped to DPbedia \cite{DBLP:journals/semweb/LehmannIJJKMHMK15} entities using DBpedia Spotlight;\footnote{See \url{https://www.dbpedia-spotlight.org/}.} these entities in turn usually correspond to Wikipedia pages, from which we can obtain the Wikipedia categories the page belongs to.  

\section{First Experiment: Assessing the Quality of Explanations Generated by Concept Induction}
\label{sec:firstExp}

In this first experiment we aimed to assess the perceived quality of machine-generated explanations (i.e., ECII’s explanations) compared to both human-generated (“gold standard”) explanations and semi-random explanations. Explanations in this study consist of brief textual descriptions of the key conceptual differences between two sets of natural images corresponding to two distinct scene categories (A and B). Note that the two image sets notionally represent the decision outputs of a binary scene classification algorithm; however, for this experiment the classifier was merely hypothetical, since the purpose of the experiment was to assess the ability of ECII to explain data differentials \emph{per se}.  The study had been reviewed and approved by New England IRB (now part of WCG IRB; protocol \#17-1325335-1) and pre-registered\footnote{See \url{https://aspredicted.org/blind.php?x=QYD_JT4} for complete study preregistration information.}  before it was conducted.

\subsection{Hypothesis}

We hypothesize that human-generated explanations will be judged most accurate by participants, followed by ECII explanations, followed by semi-random explanations.

\subsection{Method}

\subsubsection{Participants} 300 participants were recruited through Amazon Mechanical Turk using the Cloud Research platform. Participants were recruited from a listing describing the task and the compensation structure. Participants were compensated \$5 through Mechanical Turk (\$7.50 per hour prorated to the estimated time of 40 minutes required to complete the task). Based on an estimated medium effect size of f2 = 0.15 and 95\% power, we needed a sample size of at least 89 observations (i.e., unique participant judgments) per trial for estimating the parameters of the Bradley-Terry model \cite{BT52} that was used to model the data. To reach this target we aimed to collect data from 300 participants, which equates to 100 total observations for each trial. This ensured we reached the required total and still allowed for potential exclusions.

\subsubsection{Materials}

\paragraph{Image sets} A total of 45 image set pairs (A and B) were created for this study. Images were taken from the ADE20K image dataset \cite{DBLP:conf/cvpr/ZhouZPFB017,DBLP:journals/ijcv/ZhouZPXFBT19}, which includes approximately 20,000 human-curated images with scene category and object tagging. The tags (annotations) do not only indicate presence of an object but also number of such objects, occlusions, etc., but we ignored these detailed annotations for our purposes and used only the object labels, which were mapped to the Wikipedia categories as described at the end of Section \ref{sec:conceptInduction}. Each image set pair included images from two scene categories (90 scene categories in total). Each set within a pair consisted of eight images selected at random from a particular category (e.g., eight images selected randomly from the \emph{Gazebo} category). 

Also included were five “catch trial” image sets that were used to verify that participants were paying adequate attention. The images for these trials were selected in the same manner as the target trial image sets but included a different type of explanation (described below).

\paragraph{Explanations} Explanations in this study were defined as lists of up to seven concepts (i.e., strings corresponding to Wikipedia category names) that aimed to describe what was present in category A that was not present in category B (where categories A and B are the two distinct scene categories included in the image set). Concepts could be anything physically present in the images (like a computer or window), or an abstract category that fits the theme of the images (like science or entertainment).

Three types of explanations were created for each of the 45 target image sets: 1) ECII machine-generated explanations, 2) human “gold standard” explanations, and 3) semi-random machine-generated explanations.

ECII explanations were created by providing the image sets’ object tags (but not scene category tags) to the ECII algorithm. The algorithm then assessed the images and returned a rating of how well concepts matched images in category A but not category B. ECII explanations were then created by taking the seven highest rated unique concepts. ECII provides several different methods of ranking explanations. Explanations were created using rankings based on F1 scores, recall scores, precision scores, and a hybrid score. Pilot testing revealed that explanations created using F1 scores were seen as most accurate by participants, and so F1-based ECII explanations were selected for use.

Human “gold standard” explanations were created by providing the image sets (but not the object tags or scene category tags) to three human raters. Each rater independently assessed each image set and generated a list of 7 to 10 concepts that (on average) matched the images in category A but not category B. Once all raters had completed this task, gold standard explanations were created by first selecting any concept mentioned by all three raters, then concepts mentioned by two of the raters, and then finally filling the explanation with concepts randomly selected from the remaining concepts across the three raters until seven unique concepts were obtained.

Semi-random explanations were created by randomly shuffling the images in each image set to create new groupings of category A and category B that were not based on the scene category but instead contained a mixture of images from both categories. These new image sets were then provided to ECII, and explanations were generated in the same method as described above. The logic of this approach is that these explanations were likely to be viewed as more plausible than completely random explanations since they would still include concepts that are present in the images participants saw, but they would not (on average) align with one group over the other. Thus, semi-random explanations (hereafter referred to as “random” explanations) constitute a more reasonable baseline against which to assess ECII’s performance than fully random explanations, because the latter could be easily outperformed by even a low-quality explanation engine.

Catch trials included their own set of two types of explanations. One set were human explanations which were generated in the same way as the other human gold standard explanations. The other set of explanations for catch trials consisted of completely random concepts taken from a random word generator. This resulted in explanations that were very obviously inaccurate, allowing these trials to serve as an assessment of how well participants were paying attention to the task.

To standardize the presentation of explanations, all concepts in an explanation were presented in alphabetical order.

\subsubsection{Design}

Experiment  1 employed a within-subjects design in which each participant was presented with two explanations per trial and asked to choose the more accurate explanation (two-alternative forced choice design). Each participant saw all three explanation types across all trials, although only two explanation types were compared in any one trial. For each pair of image sets (A and B), the participant completed three trials comparing (1) the ECII versus human explanation; (2) the ECII versus semi-random explanation; and (3) the human versus semi-random explanation. For any given pair of image sets, a given participant completed all three comparisons (i.e., a within-subjects design).

The 45 total pairs of image sets in this study led to a total of 135 unique target trials. Because this was greater than the number of trials an individual participant was likely to be able to complete, participants were randomly assigned to 15 image sets (a total of 45 trials). Image sets were counterbalanced across participants such that all image sets received the same number of responses.

\subsubsection{Procedure}

\begin{figure*}[tb]
\includegraphics[width=\textwidth]{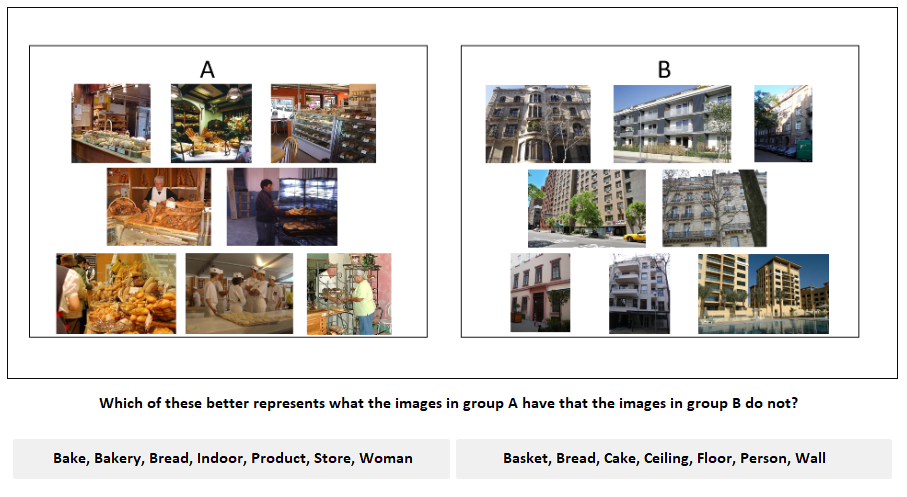}
\caption{Experiment 1 task interface, with human explanation presented on the left and ECII explanation on the right.}\label{fig:exp1}
\end{figure*}

After providing consent, participants completed brief training on the task, including instructions about how concepts and explanations were defined in this study. Participants then began completing trials. The 50 trials (45 assigned targets and 5 catch trials) were presented in a random order. Figure \ref{fig:exp1} shows what the stimuli presentation and response options looked like to participants.

\subsection{Results}

\begin{figure}[tb]
\includegraphics[width=\columnwidth]{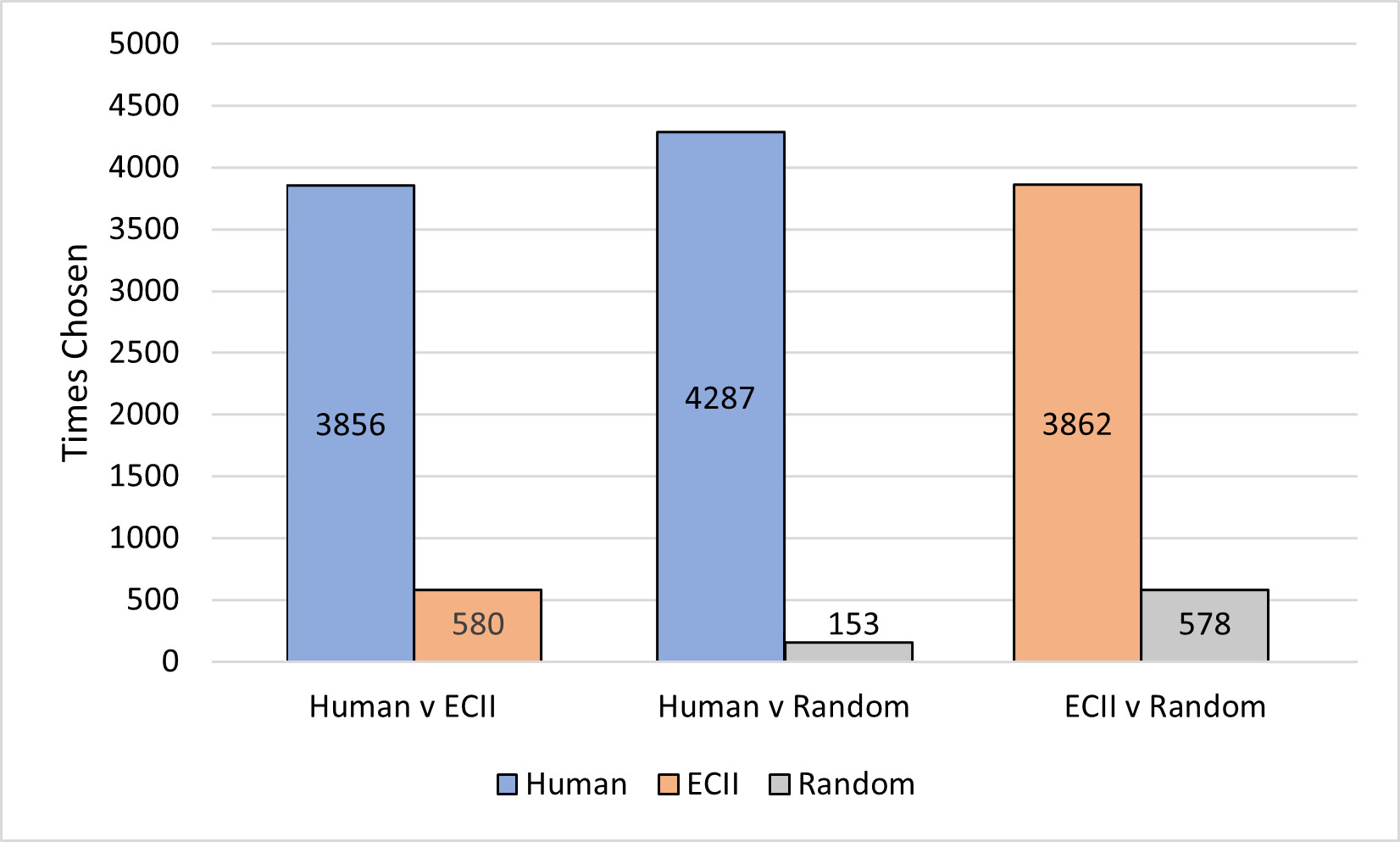}
\caption{Number of times participants chose different explanation types in Experiment 1.}\label{fig:exp1res}
\end{figure}

Prior to analysis, participant responses to the catch trials were assessed. Participants who failed more than one catch trial were excluded from analysis. While the vast majority of participants did not fail any catch trials (N=295) and one participant failed exactly one trial, four participants failed more (two failed two trials, and two failed three trials). These four participants were excluded from all analyses, leaving a total of 296 participants included in analyses. 
Across all image sets, human explanations were overwhelmingly chosen over ECII explanations (3856 vs 580; i.e., 87\% of the time) and over random explanations (4287 vs 153; i.e., 97\% of the time), and ECII explanations were chosen overwhelmingly over random explanations (3862 vs 578; i.e., 87\% of the time). See Figure \ref{fig:exp1res}.

Participants’ pairwise judgments were used in a Bradley-Terry analysis to obtain “ability scores” for each type of explanation, where ability scores in this analysis provide a metric of how much an explanation type was preferred by participants. Ability scores were calculated for each of the 45 image sets. Analysis of these ability scores revealed that human explanations had the highest ability scores (M = 5.32, SD = 3.87), followed by ECII explanations (M = 3.08, SD = 3.99), F(2) = 30.37, $p < 0.001$, $\eta^2 = 0.315$). Random explanations were used as the comparison point in the Bradley-Terry analysis, and thus were set to 0 (with the ability scores for human and ECII explanations indicating how much they were preferred compared to the random explanations). A Tukey’s HSD test revealed that differences in the ability scores of human vs. random explanations and ECII vs. random explanations were both significant at $p < 0.001$, while differences in the ability scores of human vs. ECII explanations were significant at $p = 0.004$. See Table \ref{tab:exp1as} for the individual human and ECII ability scores for each image set pair in Experiment 1.

\begin{table*}[p]
    \centering
\begin{tabular}{l|rr|ccc}
Image Set                            & H. Ability & E. Ability & H. v E. Wins & H. v R Wins & E v R Wins \\
\hline
Set 1: Bedroom v Park                & 4.58       & 2.04       & 90 – 6       & 94 – 2      & 86 – 10    \\
Set 2: Living Room v Parking Lot     & 7.40       & 3.91       & 97 – 2       & 98 – 1      & 98 – 1     \\
Set 3: Office v Playground           & 6.09       & 4.80       & 76 – 20      & 95 – 1      & 96 – 0     \\
Set 4: Airport v Amusement Park      & 2.60       & 0.09       & 93 – 4       & 87 – 10     & 54 – 43    \\
Set 5: Bathroom v Art Studio         & 5.10       & 3.66       & 78 – 18      & 95 – 1      & 94 – 2     \\
\hline
Set 6: Beauty Salon v Forest Path    & 4.25       & 3.60       & 65 – 33      & 96 – 2      & 96 – 2     \\
Set 7: Bookstore v Child Room        & 5.42       & 2.80       & 91 – 6       & 96 – 1      & 92 – 5     \\
Set 8: Hotel Room v Cockpit          & 5.35       & 4.17       & 78 – 22      & 98 – 2      & 100 – 0    \\
Set 9: Shoe Store v Alcove           & 4.12       & 3.09       & 75 – 26      & 99 – 2      & 97 – 4     \\
Set 10: Alley v Wet Bar              & 4.94       & 1.19       & 98 – 2       & 99 – 1      & 77 – 23    \\
\hline
Set 11: Closet v Construction Site   & 7.76       & 4.61       & 93 – 4       & 97 – 0      & 96 – 1     \\
Set 12: Gazebo v Bowling Alley       & 5.31       & 2.49       & 93 – 5       & 97 – 1      & 91 – 7     \\
Set 13: Garage v Hallway             & 28.74      & 27.86      & 70 – 29      & 99 – 0      & 99 – 0     \\
Set 14: Laundromat v Pantry          & 6.77       & 3.54       & 98 – 4       & 102 – 0     & 99 – 3     \\
Set 15: Conference Room v Waterfall  & 6.97       & 3.49       & 95 – 3       & 98 – 0      & 95 – 3     \\
\hline
Set 16: Home Office v Bow Window     & 4.29       & 1.93       & 93 – 7       & 97 – 3      & 89 – 11    \\
Set 17: Dining Room v Kitchen        & 1.77       & 1.64       & 54 – 44      & 81 – 16     & 83 – 14    \\
Set 18: Fast Food v Office Building  & 4.09       & 1.26       & 91 – 6       & 96 – 1      & 75 – 22    \\
Set 19: Jacuzzi v Greenhouse         & 5.83       & 2.24       & 98 – 3       & 101 – 0     & 91 – 10    \\
Set 20: Gymnasium v Corridor         & 3.23       & -0.12      & 93 – 4       & 94 – 3      & 45 – 52    \\
\hline
Set 21: Bus v Broadleaf Forest       & 6.75       & 3.22       & 99 – 2       & 100 – 1     & 98 – 3     \\
Set 22: Casino v Arrival Gate        & 6.17       & 3.55       & 95 – 5       & 98 – 2      & 99 – 1     \\
Set 23: Library v Gas Station        & 4.61       & 3.40       & 77 – 23      & 94 – 1      & 92 – 3     \\
Set 24: Valley v Yard                & 3.74       & -1.17      & 98 – 0       & 95 – 3      & 24 – 74    \\
Set 25: Mountain v Coast             & 0.64       & 0.32       & 58 – 40      & 63 – 35     & 58 – 40    \\
\hline
Set 26: Dinette Vehicle v Farm Field & 4.81       & 2.92       & 90 – 11      & 98 – 3      & 98 – 3     \\
Set 27: Poolroom v Driveway          & 1.61       & 2.78       & 23 – 75      & 82 – 16     & 92 – 6     \\
Set 28: Bridge v Auditorium          & 5.15       & 3.04       & 87 – 10      & 96 – 1      & 94 – 3     \\
Set 29: Museum v Youth Hostel        & 1.85       & -0.10      & 89 – 11      & 85 – 15     & 49 – 51    \\
Set 30: Supermarket v Restaurant     & 6.16       & 3.97       & 88 – 9       & 96 – 1      & 96 – 1     \\
\hline
Set 31: Classroom v Archive          & 3.97       & 2.27       & 84 – 14      & 95 – 3      & 90 – 8     \\
Set 32: Dentist Office v Ballroom    & 3.79       & 1.25       & 88 – 6       & 95 – 3      & 77 – 21    \\
Set 33: Lighthouse v River           & 4.51       & 2.08       & 91 – 7       & 97 – 1      & 88 – 10    \\
Set 34: Creek v Basement             & 6.87       & 2.94       & 97 – 1       & 97 – 1      & 94 – 4     \\
Set 35: Building Facade v Ocean      & 4.32       & 2.00       & 90 – 8       & 96 – 2      & 87 – 11    \\
\hline
Set 36: Courthouse v Parking Garage  & 2.28       & -0.45      & 90 – 8       & 91 – 7      & 36 – 62    \\
Set 37: Balcony v Skyscraper         & 5.39       & 2.69       & 96 – 7       & 103 – 0     & 96 – 7     \\
Set 38: Game Room v Waiting Room     & 4.70       & 3.83       & 69 – 30      & 99 – 0      & 96 – 3     \\
Set 39: Landing Deck v Window Seat   & 4.82       & 2.28       & 93 – 6       & 97 – 2      & 91 – 8     \\
Set 40: Bar v Warehouse              & 4.81       & 2.87       & 86 – 11      & 95 – 2      & 93 – 4     \\
\hline
Set 41: Bakery v Apartment Building  & 5.41       & 3.64       & 86 – 14      & 99 – 1      & 98 – 2     \\
Set 42: Needleleaf Forest v Playroom & 7.03       & 4.70       & 92 – 9       & 101 – 0     & 100 – 1    \\
Set 43: Outdoor Window v Roundabout  & 4.58       & 1.10       & 98 – 2       & 98 – 2      & 76 – 24    \\
Set 44: Reception v Golf Course      & 4.03       & 1.73       & 94 – 7       & 97 – 4      & 88 – 13    \\
Set 45: Staircase v Plaza            & 6.46       & 4.78       & 85 – 16      & 101 – 0     & 100 – 1   
\end{tabular}    
\caption{Ability Scores and Number of Wins for Human (H.), ECII (E.), and Random (R.) Explanations. Note that random explanations were set as the reference point in the Bradley-Terry analysis and so their ability scores were always equal to 0, and thus are not displayed here.}
    \label{tab:exp1as}
\end{table*}

\subsection{Discussion}

Analysis of the results of Experiment 1 provide evidence to support our main hypothesis. Participants found the human-generated explanations to be the most accurate at describing the difference in the image sets, followed by the ECII explanations, and the semi-random explanations proving to be the least accurate. This provides support for the value of ECII explanations. It is not surprising that explanations produced by the ECII algorithm are of lower quality than human-generated explanations at this stage of development. However, the ECII explanations do contain notable explanatory power, suggesting that ECII explanations can be useful. It should be noted also that there was some variability in ECII’s performance across the image sets. For some image sets the ECII explanations were chosen relatively more often than in others, in one case even being chosen more frequently than the human explanation, while in other image sets it was chosen less often than the random explanation. This suggests there is certainly still room for improvement in ECII explanations, but that on average there is promising evidence that ECII can produce explanations that accurately describe the differences between two groups of data.

\section{Second Experiment: Identifying Errors Made by a Machine Learning Classifier}
\label{sec:secondExp}

In Experiment 1 the quality of ECII-generated explanations was (subjectively) judged by human participants via comparison with human-generated (“gold standard”) and random-generated explanations. Our second experiment (Experiment 2) sought to obtain an objective measure of the utility of ECII’s explanations in helping human users evaluate the decisions of an actual ML system (more precisely, a logistic regression algorithm that classified images into scene categories based on semantic tags of objects present in each image). Specifically, the goal of this second experiment was to test how well ECII explanations (compared to human “gold standard” explanations) helped human participants identify the errors made by the AI. Note that a key difference between this experiment and Experiment 1 is that the explanations in Experiment 1 were of a (hypothetical) ML's \emph{decisions}, whereas the explanations in Experiment 2 were of a (real) ML's decision \emph{errors}. The study had been reviewed and approved by New England IRB (now part of WCG IRB; protocol \#17-1325335-1) and pre-registered\footnote{See \url{https://aspredicted.org/blind.php?x=4FH_1X4} for complete study preregistration information.} before it was conducted.

\subsection{Hypothesis}

We hypothesized that participants would be able to match human generated explanations to the correct image set more frequently than for ECII explanations, and that participants would be better than chance for both types of explanations.

\subsection{Method}

\subsubsection{Participants}

Amazon Mechanical Turk participants were recruited using the Cloud Research platform. Participants were recruited from a listing on Mechanical Turk describing the task and the compensation structure. Participants were compensated \$5 through Mechanical Turk (\$7.50 per hour prorated to the estimated time of 40 minutes required to complete the task). Following recommendations by \cite{Rou14}, we initially planned to collect data from 100 participants and compute Bayes Factors for our effects of interest. If the Bayes Factors was not conclusive, then we would continue to collect data in groups of 50 participants, stopping to analyze the data after each new cohort, ceasing data collection if we reached a sample size of 250. We did not need to collect more than the initial 100 responses.

\subsubsection{Materials}

\paragraph{Image sets}

A total of 16 image sets were created for this study. Images were taken from the Google Open Images database, which includes object tagging and scene category tags. The annotated tags in these images are crowd-generated and thus noisy. Limited manual curation was therefore performed as part of the image selection process: for example, images with very few object tags (5 or fewer) were removed from the dataset, as images with too few labels would not provide the classifier with enough context to make a classification. Each training set included images from both the target category and images from multiple non-target categories. All images from the target category with sufficient object tags were included. An equal number of images were drawn randomly from non-target categories to provide a balanced dataset for each image set. We utilized the following 16 scene categories: bakeries, bathrooms, bedrooms, bridges, cafes, classrooms, dining rooms, gazebos, greenhouses, kitchens, lobbies, offices, pantries, parks, parking lots, and libraries.

Selected images were fed to a logistic regression classifier to classify scene images into target / non-target categories based on their tags. We utilized 10-fold cross validation to train and test the classifier. The input stimuli were represented as binary object vectors indicating the presence or absence of each object tag in that image. The vector space was generated by taking a set of all the objects present in the image dataset for a target scene category (e.g., all object tags present in the kitchen vs. non-kitchen image set). The classifier then outputs a binary decision as to whether each image is part of the target set or not. These classifications were then grouped into four categories based on ground truth scene categories: true positives (TP), true negatives (TN), false positives (FP), and false negatives (FN). Up to nine images from each grouping were included in final image sets displayed to participants, randomly selected from the full set of images in each error group.
The same five catch trial image sets from Experiment 1 were also included in Experiment 2.

\paragraph{Explanations}

Explanation generation in Experiment 2 followed the same process as in Experiment 1, with the exception that only two types of explanations were used for Experiment 2: 1) ECII (machine-generated) explanations and 2) human “gold standard” explanations. Due to the particular design of the study, the semi-random explanations would not have provided additional information and therefore were not used.

ECII explanations were again created by providing the image set’s object tags (but not scene category tags) to the ECII algorithm, which provided ratings of concepts in the same manner as in Experiment 1. Object tags from all images in a set were provided to ECII, not only those in the subset of images displayed to participants. Explanations were generated for four image sets for each scene category: FP vs TN, TP vs FN, TP vs FP, and FN vs TN. These four comparisons were chosen because they were deemed most likely to have explainable differences. For example, we reasoned that FP scenes might contain a restricted set of defining elements that distinguish them from TN scenes – e.g., FP kitchen images might be more likely to contain kitchen-relevant features that TN kitchens do not have, such as images of offices that contain microwave ovens). In contrast, FP scenes could contain a wide variety of elements that distinguish them from TP scenes (e.g., a FP kitchen could contain an oven in a warehouse filled with myriad other objects) or from FN scenes (e.g., a FP kitchen and FN kitchen could both contain a large set of kitchen-irrelevant features that are hard to capture concisely).

\begin{figure}[tb]
\includegraphics[width=\columnwidth]{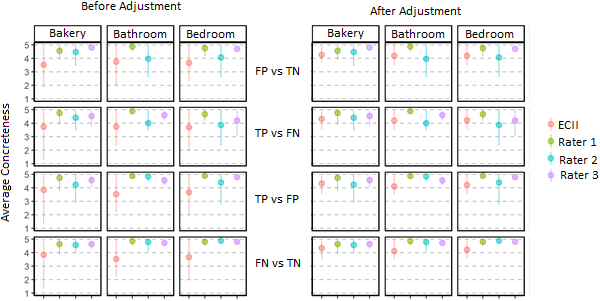}
\caption{Average concreteness in three image groups for ECII and human concepts before and after adjustment.}\label{fig:exp2conc}
\end{figure}

Explanations in Experiment 1 tended to be more abstract, featuring thematic or high-level category concepts, unlike the very concrete explanations created by human raters, which tended to feature specific objects in the image sets. In order to equate the concreteness of the ECII and human explanations in Experiment 2, we conducted an analysis of the concreteness of the terms used in human explanations (described below), using concreteness ratings provided by \cite{BWK14}. We assessed the concreteness of each concept provided by human raters and in the ECII explanations and confirmed that ECII concepts were noticeably less concrete than the concepts provided by all three human raters. In order to adjust the concreteness of ECII explanations, we eliminated all concepts provided by ECII with a concreteness score of less than 3.5 (See Figure \ref{fig:exp2conc} for an example of the concreteness of ECII explanations before and after adjustment, compared to the concreteness of the concepts provided by the three human raters). Explanations were then created by taking the seven highest rated concepts from the filtered set returned by ECII. 

\begin{figure}[tb]
\includegraphics[width=\columnwidth]{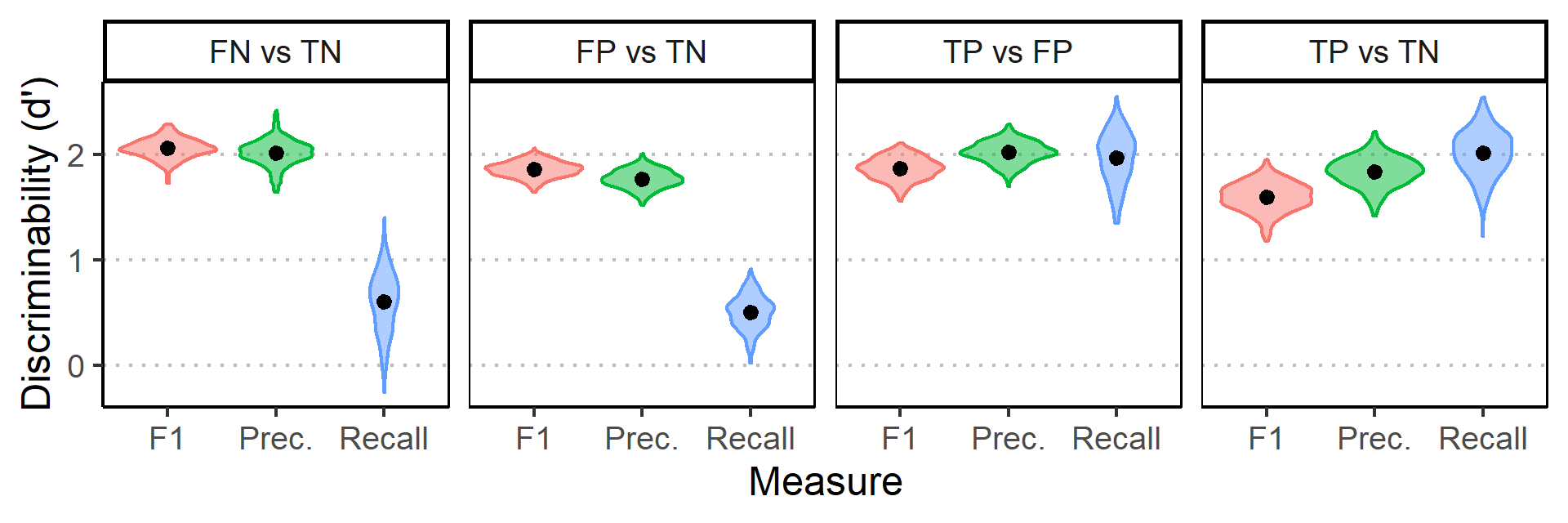}
\caption{Discriminability as a function of comparison (noted in the panel titles) and explanation type (maximized with respect to F1, precision, or recall) for our pilot study. Violins indicate the distribution of participants’ d’ values. Black dots indicate group means.}\label{fig:exp2pilot}
\end{figure}

Explanations generated by ECII can be ranked with respect to several criteria; to determine which criteria would be most helpful for human participants, we ran a pilot study in which we compared the effectiveness of explanations maximized with respect to F1, precision, and recall. Discriminability from this pilot study is plotted in Figure \ref{fig:exp2pilot}. Overall, we found that all three explanation types generated similar discriminability for our “TP vs. FP” and “TP vs. TN” comparisons, and that explanations maximized with respect to F1 and precision yielded similar discriminability for the remaining two comparisons. However, only F1 generated higher discriminability than recall-based explanations for the “FN vs. TN” comparison, whereas precision discriminability was not convincingly better than that of recall (i.e., BF $<$ 3). We therefore elected to use explanations that were generated with respect to F1 scores in Experiment 2.

Human “gold standard” explanations were again created by providing the image sets (but not object tags or scene category tags) to three human raters. Raters again independently generated concepts for the same four comparisons for each scene category that ECII did. Gold standard explanations were created by pooling the concepts humans provided following the same procedure as in Experiment 1.

Catch trials utilized the same explanations as in Experiment~1, although the randomly generated explanations were not included due to the design of Experiment 2.

Because participants did not need to compare two different explanations in Experiment 2, concepts were not presented in alphabetical order, unlike in Experiment 1.

\subsubsection{Design}

Experiment 2 utilized a fully crossed within-subjects design with 2 (explanation type: ECII, human) $\times$ 4 (comparison sets: FP vs TN, TP vs FN, TP vs FP, FN vs TN) conditions for a total of 8 conditions. Participants were randomly assigned to 8 (out of 16) scene categories. For each scene category, participants saw all four comparisons. For half of the categories, they saw the human explanations, and for the other half they saw the ECII explanation. This resulted in a total of 32 target trials for each participant (4 in each explanation type $\times$ comparison set condition).

\subsubsection{Procedure}

\begin{figure*}[tb]
\includegraphics[width=\textwidth]{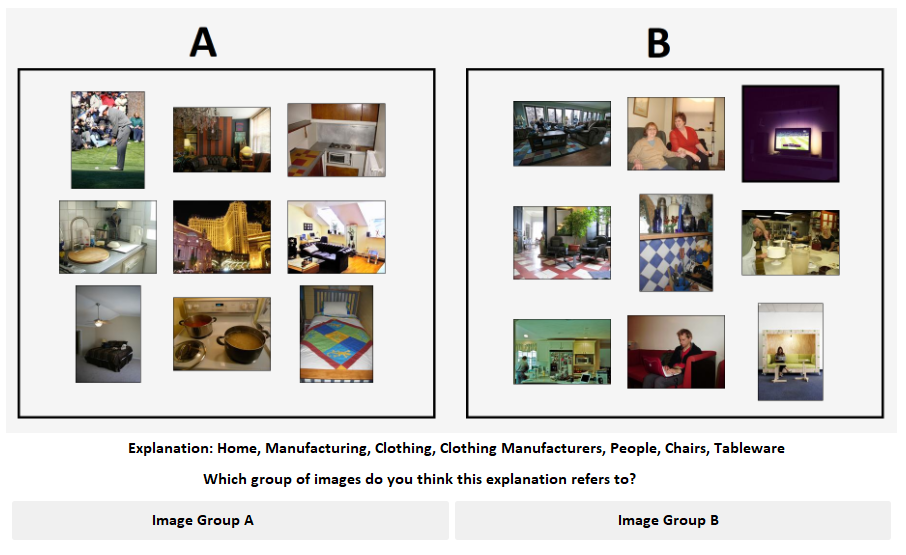}
\caption{Experiment 2 task interface, with ECII explanation referring to image group B.}\label{fig:exp2}
\end{figure*}

After providing consent, participants completed brief training on the task, including instructions about how concepts and explanations were defined in this study. Participants then began completing trials. The 37 trials (32 assigned targets and 5 catch trials) were presented in a random order. Figure \ref{fig:exp2} shows what the stimuli presentation and response options looked like to participants.

\subsection{Results}

We analyzed choice data from Experiment 2 with a Bayesian hierarchical signal-detection model (SDT; \cite{GS66}). In its simplest form, SDT parameterizes binary choices as a function of discriminability (i.e., $d'$) and bias (i.e., $c$). Discriminability is the distance between a posited mental Gaussian distribution of noise and a Gaussian distribution of signal. Bias is the degree to which an individual’s decision threshold varies from ideal placement (i.e., the point at which the signal and noise distributions are equally likely); negative or positive bias indicates a tendency to respond with one choice over another. This separation of discriminability from bias ensures that individuals are not misjudged as accurate when they generate one response more often than the other (e.g., consider a student who achieves 50\% accuracy on a balanced true-false exam by always responding “true”). 

Our SDT model was estimated in Stan probabilistic language \cite{CGH+17} via the “brms” package \cite{DBLP:journals/jstatsoft/Burkner21} in R statistical software. We estimated population-level coefficients for discriminability and bias for each comparison and explanation type (i.e., ECII versus humans), as well as participant- and scene-level effects for discriminability and bias. Population-level coefficients were assessed with Bayes factors, which were estimated via Savage-Dickey ratios \cite{WAGENMAKERS2010158}. Following recommendations from \cite{Jef61}, we considered evidence as convincing once it supported one hypothesis over another by a 3:1 ratio.

Because our analyses were rooted in Bayesian estimation, we took advantage of optional stopping during data collection \cite{Rou14}. Our primary interest was in evaluating whether (a) human- and ECII-generated explanations facilitated set discrimination to different degrees and (b) whether human- and ECII-generated explanations yielded non-zero discriminability. We therefore elected to initially collect data from 100 individuals, and then collect data in cohorts of 50 participants either until we (a) found convincing evidence in favor of either the null or alternative hypothesis (i.e., a Bayes factor either $\geq 3$ or $\leq 0.33$) or else (b) collected data from 250 people total. We ultimately stopped data collection after 100 participants.

\begin{figure}[tb]
\includegraphics[width=\columnwidth]{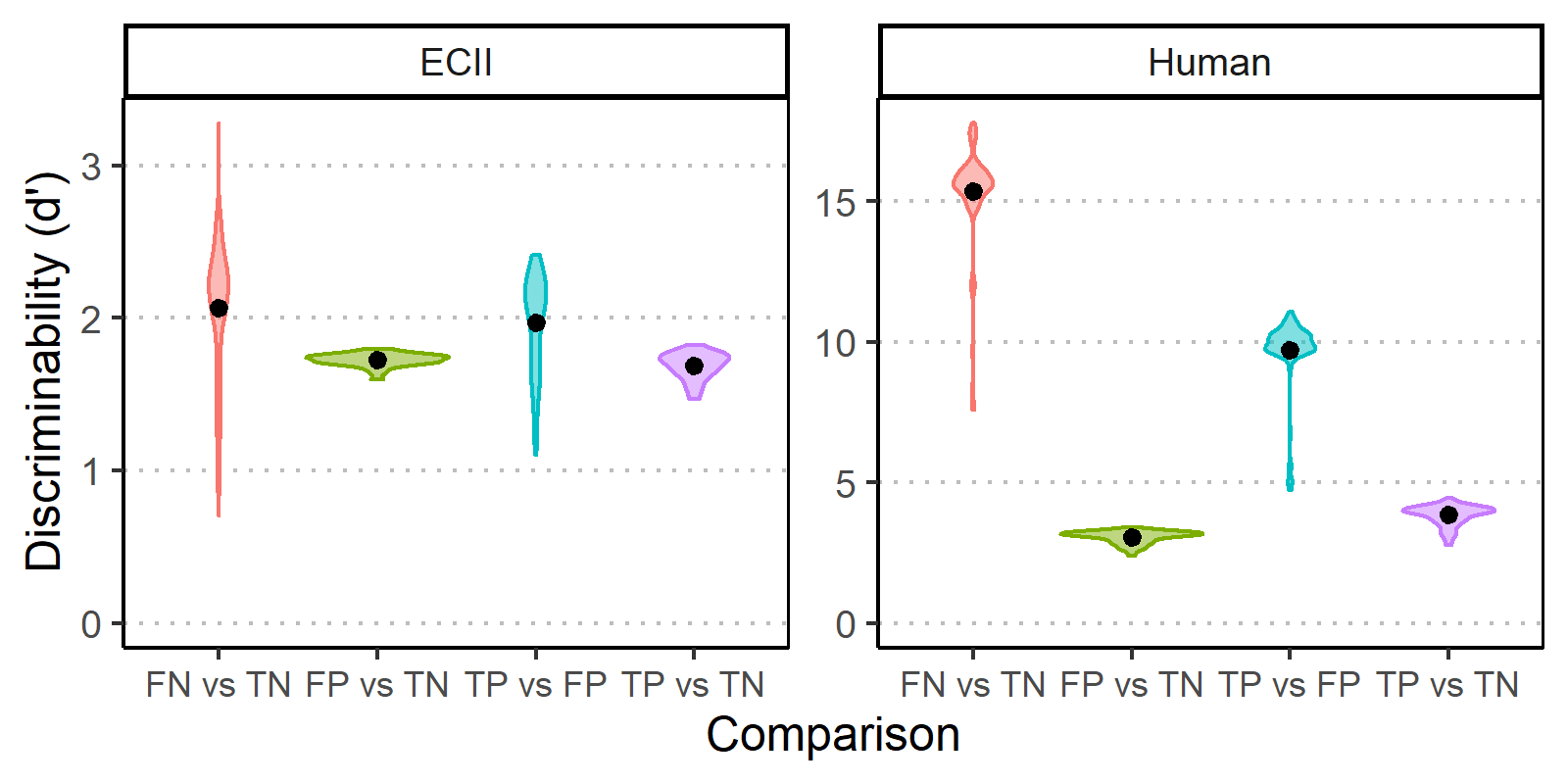}
\caption{Discriminability as a function of comparison (noted on the abscissa) and explanation type (noted in the panel titles) for Experiment 2. Violins indicate the distribution of participants’ $d'$ values. Black dots indicate group means. Note the different scales in the ordinates.}\label{fig:exp2disc}
\end{figure}

Estimates of participants’ discriminability are plotted in Figure~\ref{fig:exp2disc}. Population-level estimates of discriminability and bias as a function of explanation type are reported in Table \ref{tab:exp2disc}; differential values between humans and ECII are reported in Table \ref{tab:exp2diffdisc}. Human-generated explanations yielded higher discriminability for all four of our comparisons. For both human-generated and ECII-generated explanations, we found evidence for non-zero $d'$. Thus, while ECII is not as effective as humans at generating helpful descriptions of set differences, it is nonetheless still an effective intervention across all comparisons that we evaluated. Additionally, both sets of explanations yielded a bias toward responding with “Set A” (but note that correct responses were balanced across “Set A” and “Set B”), and this bias was consistently stronger for human-generated explanations. 

\begin{table}[tb]
\begin{center}\footnotesize
\begin{tabular}{c|ccccc}
Source            & Comparison & $d'$    & BF${}_{10}$        & $c$     & BF${}_{10}$         \\
\hline
ECII              & FN vs. TN  & 2.06  & 21.04       & -0.65 & 1.91         \\
                & FP vs. TN  & 1.74  & 4.82 × 1015 & -0.75 & 1.73 × 1012  \\
                  & TP vs. FP  & 1.97  & 2.20 × 1013 & -0.73 & 38.30        \\
                  & TP vs. TN  & 1.68  & 169.21      & -0.78 & 55.97        \\
\hline
Human             & FN vs. TN  & 15.31 & 1.71 × 105  & -4.74 & 5.18 × 108   \\
                 & FP vs. TN  & 3.05  & 4.37 × 1024 & -1.46 & 3.60 × 10109 \\
                  & TP vs. FP  & 9.69  & 4.51 × 106  & -2.74 & 5.28 × 1017  \\
                  & TP vs. TN  & 3.83  & 4.16 × 1017 & -1.66 & 2.20 × 10103
\end{tabular}
\end{center}
\caption{Estimates of Discriminability ($d'$) and Bias ($c$) for ECII- and Human-Generated Terms Across Four Comparison Types. Note that columns, from left to right, indicate the source of explanations, the comparison being evaluated, estimated discriminability, the corresponding Bayes factor for discriminability, estimated bias, and the corresponding Bayes factor for bias. Negative/positive values of bias indicate a propensity to choose Set A/B.}\label{tab:exp2disc}
\end{table}

\begin{table}[tb]
\begin{center}
\begin{tabular}{c|rrrr}
Comparison & $d'_{\text{diff}}$ & BF${}_{10}$   & $c_{\text{diff}}$ & BF${}_{10}$   \\
\hline
FN vs. TN  & 13.25  & 322.01 & -4.10 & 493.89 \\
FP vs. TN  & 1.33   & 210.39 & -0.71 & 260.20 \\
TP vs. FP  & 7.72   & 210.48 & -2.01 & 826.99 \\
TP vs. TN  & 2.15   & 62.59  & -0.88 & 4.49  
\end{tabular}
\end{center}
\caption{Estimates of Differential Discriminability ($d'_{\text{diff}}$) and Bias ($c_{\text{diff}}$) between ECII- and Human-Generated Terms Across Four Comparison Types. Note that columns, from left to right, indicate the comparison being evaluated, the differential estimated discriminability (i.e., $d'_\text{human} - d'_\text{ECII}$), the corresponding Bayes factor, differential estimated bias (i.e., $c_\text{human} - c_\text{ECII}$), and the corresponding Bayes factor.}\label{tab:exp2diffdisc}
\end{table}

\subsection{Discussion}

Experiment 2 suggests that ECII can generate sensible explanations that help humans to distinguish misclassified scenes from correctly classified ones. Explanations generated by humans were even more helpful, which is not entirely unexpected, but our findings nonetheless suggest that ECII may be deployed to effectively facilitate human analysts’ error detection for AI classification. Another notable result from Experiment 2 is that human explanations generated more biased responding in addition to more discriminative responses; while we did not anticipate such an outcome, one plausible explanation is that participants were randomly responding more frequently when viewing ECII explanations (i.e., because ECII was not as interpretable, despite our attempts to increase the concreteness of its terms), whereas human explanations encouraged more systematic (if also biased) responding. Overall, our findings suggest that ECII could serve as an effective teammate for facilitating human analysts’ detection of misclassified entities.

One question that we have not addressed directly in Experiment~2 is how humans detect misclassified sets. For example, did participants notice that the error images include/exclude certain objects that the correct images do/do not contain? Future research on this topic could employ eye-tracking methods to evaluate which particular scenes, or specific features thereof, people tend to assess prior to making their choices.

\section{Related Work}
\label{sec:related}

Despite significant recent research efforts regarding approaches to XAI, this area of research is still in its infancy and in need of new ideas. Explanation methods so far can be divided into two broad categories, one focusing on local explanations, and one focusing on global explanations. Local explanation algorithms seek understanding of individual predictions, while global explanation algorithms seeks understanding of the overall behavior of the deep learning model. 

Local explanation approaches can in turn be divided into 4 broad categories, a) those focusing on feature importance, b) those using saliency maps, c) those that are prototypes or examples based, and d) those based on counterfactuals. As the name suggests, feature importance algorithms attempt to identify which input features are more important for the prediction. Popular feature importance algorithms are LIME \cite{ribeiro16} and SHAP\cite{NIPS2017_7062}. State-of-the-art saliency map based algorithms are \cite{DBLP:journals/corr/SelvarajuDVCPB16, DBLP:journals/corr/abs-1908-01224}. Prototype/example based algorithms seek to explain  individual decisions in terms of other examples (whether artificial or based on real data). Popular algorithms in this category make use of so-called influence functions \cite{DBLP:conf/icml/KohL17}, and a corresponding survey on this can be found in \cite{DBLP:series/lncs/NguyenYC19}. A popular counterfactuals based algorithm was provided in \cite{DBLP:conf/icml/GoyalWEBPL19}, and an overview of counterfactual explanations can be found in \cite{DBLP:journals/corr/abs-2010-10596}. 

Global explanation approaches can be categorized into 3 broad categories, a) those that are based on collections of local explanations like SP-LIME \cite{ribeiro16}, b) those that are representation based like TCAV \cite{DBLP:conf/icml/KimWGCWVS18}, and c) those based on model distillation like \cite{tan2019learning, DBLP:conf/aiia/FrosstH17}). State of the art algorithms have issues regarding faithfulness \cite{DBLP:conf/nips/AdebayoGMGHK18}, stability \cite{DBLP:journals/corr/abs-1806-08049, DBLP:conf/cvpr/BansalAN20} and fragility \cite{DBLP:series/lncs/KindermansHAASDEK19, DBLP:conf/aaai/GhorbaniAZ19}.

The use of background or domain knowledge to produce or enhance explanations is gaining attention very recently. Domain knowledge comes in the form of simple concept tags, knowledge graphs (essentially data organized in graphs) or ontologies (knowledge bases using formal logic with formal semantics). For the use of concepts, one of the pioneering contributions was by Been Kim et. al.{} who showed the value of simple concepts to produce human understandable explanations \cite{DBLP:conf/icml/KimWGCWVS18}. Knowledge graph were used, for example, to explain individual decisions in the context of stock trend predictions \cite{DBLP:conf/www/DengZZCPC19}. Ontologies have been used in several lines of research for explanation. In the transfer learning domain, they were used to obtain an understanding which features are worth transferring or not \cite{DBLP:journals/corr/abs-1901-08547}. The publication \cite{DBLP:conf/fat/PaniguttiPP20} introduced Doctor XAI as a model-agnostic explainer that focuses on explaining medical diagnosis predictions. They show how exploiting temporal dimensions in the data together with domain knowledge encoded as a medical ontology improves the quality of mined explanations. Trepan-reloaded \cite{DBLP:journals/ai/ConfalonieriWBM21} is an extension of the original Trepan \cite{DBLP:conf/nips/CravenS95} algorithm, and the authors demonstrate the use of ontologies for producing explanations that are more understandable by humans. Though, as we have just seen, there exists XAI research that makes use of domain knowledge, concept induction for making use of background knowledge is not yet a prevalent method. We are only aware of our own preliminary investigations \cite{SarkerXDRH17} and \cite{DBLP:conf/semco/ProckoEOR22} that make use of concept induction.

Among the many approaches to XAI, some explanations are generated for end users, and some explanations are meant for system developers, but in any case it remains an important criterion that an explanation makes sense to a human. To evaluate which explanations are better than others in this respect, different methods are being proposed \cite{DBLP:conf/chi/Poursabzi-Sangdeh21, gacto2011interpretability, carvalho2019machine, DBLP:journals/corr/abs-1812-04608}. And while the different approaches are being evaluated, it is also important to keep in mind the variability of human understanding \cite{DBLP:journals/corr/abs-1802-00682}.


\section{Conclusions and Future Work}
\label{sec:future} \label{sec:conc}

The results reported herein clearly indicate that concept induction is an approach to explanation generation over background knowledge that produces explanations that are meaningful at a human level. But of course, the studies presented herein are only first steps on a longer journey towards making the approach useful in practice. We discuss several lines of investigation that can follow up on the presented results.

It appears a reasonable hypothesis that more expressive background knowledge -- e.g. a highly axiomatized ontology -- could yield even better and more fine-grained explanations. In fact, explanations could even include individuals in the form of nominal classes, i.e. reasoning would be over the schema (ontology) and a corresponding knowledge graph to derive explanations. A key technical hurdle remains for this, though: While ECII scales to the task, its heuristic relies on an ontology that is merely a class hierarchy, and other algorithms like those underlying DL-Learner do not scale sufficiently. Improved algorithms and systems -- heuristic or not -- for concept induction therefore need developing, before expressive background knowledge can be used at scale.

Another rather obvious avenue is in exploring concept induction for explanation generation in other settings, and in particular for deep learning systems that are inherently black boxes. One the one hand, an analysis of type 1 and type 2 errors by a trained network could lead to the identification of commonalities of inputs that produce errors, which could then be corrected, e.g., by including more training samples of the error-producing type. Another possible line of research is the use of concept induction to explain hidden layer activation patterns.

While concept induction is the key mechanism underlying our study, it appears to us that the choice of background knowledge is probably rather critical. Indeed, we also experimented with other hierarchies, e.g., SUMO \cite{DBLP:conf/kgswc/SarkerSHZNMJRA20}, but the results were not as convincing. To advance, a systematic exploration of the effects of different types of background knowledge appears to be in order.

Besides the above mentioned immediate next steps, of course there also remains the broader vision for our work: The engineering of a human-machine data analysis support system that uses background information to provide explanations to a human analyst. 

\section*{Acknowledgement}
\label{sec:acknowldge}

This research was funded by the U.S. Government via contract FA8649-20-9-9045 to Kairos Research. The views and conclusions contained in this document are those of the authors and should not be interpreted as representing the official policies, either expressed or implied, of the U.S. Government.



\bibliographystyle{abbrv}
\bibliography{refs}

\end{document}